%% file: GPTNSS.tex
\documentclass{article}

\usepackage{microtype}
\usepackage{graphicx}
\usepackage{subfigure}
\usepackage{booktabs} 
\usepackage{multirow}
\usepackage{hyperref}

\usepackage[accepted]{icml2024}

\usepackage{amsmath}
\usepackage{amssymb}
\usepackage{mathtools}
\usepackage{amsthm}
\usepackage{enumitem}
\usepackage[frozencache,cachedir=.]{minted}

\usepackage{algpseudocodex}

\usepackage{makecell}

\makeatletter
\newcommand*\bigcdot{\mathpalette\bigcdot@{.5}}
\newcommand*\bigcdot@[2]{\mathbin{\vcenter{\hbox{\scalebox{#2}{$\m@th#1\bullet$}}}}}
\makeatother

\usepackage[capitalize,noabbrev]{cleveref}

\theoremstyle{plain}

\theoremstyle{definition}

\theoremstyle{remark}

\usepackage[textsize=tiny]{todonotes}
\usepackage{mathrsfs}
\usepackage{threeparttable}
\icmltitlerunning{tnGPS: Discovering Unknown Tensor Network Structure Search Algorithms via Large Language Models}

\begin{document}

\twocolumn[

\icmltitle{tnGPS: Discovering Unknown Tensor Network Structure Search Algorithms \\ via Large Language Models (LLMs)}



\icmlsetsymbol{equal}{*}

\begin{icmlauthorlist}
\icmlauthor{Junhua Zeng}{GUT,equal}
\icmlauthor{Chao Li}{riken,equal}
\icmlauthor{Zhun Sun}{sch}
\icmlauthor{Qibin Zhao}{riken}
\icmlauthor{Guoxu Zhou}{GUT,addi}
\end{icmlauthorlist}

\icmlaffiliation{GUT}{School of Automation, Guangdong University of Technology, Guangzhou, China}
\icmlaffiliation{riken}{RIKEN Center for Advanced Intelligence Project (RIKEN-AIP), Tokyo, Japan}
\icmlaffiliation{addi}{Key Laboratory of Intelligent Detection and the Internet of Things in Manufacturing, Ministry of Education, Guangzhou, China}
\icmlaffiliation{sch}{Tencent Inc., Shenzhen, China}
\icmlcorrespondingauthor{Guoxu Zhou}{gx.zhou@gdut.edu.cn}

\icmlkeywords{tensor, tensor network, model selection, large language models}

\vskip 0.3in
]



\printAffiliationsAndNotice{\icmlEqualContribution} 

\input{sections/00_abstract}

\section{Introduction}
\label{Section:Intro}
Tensor networks (TNs) are powerful methods that have proven to be highly beneficial in machine learning and various other fields~\cite{markov2008simulating,anandkumar2014tensor,orus2014practical,novikov2015tensorizing,cichocki2016tensor,stoudenmire2016supervised,cichocki2017tensor,orus2019tensor,glasser2019expressive,kossaifi2020tensor,miller2021tensor,richter2021solving,miller2021tensor,haghshenas2022variational}. 
Their effectiveness lies in the ability to decompose extremely high-dimensional problems into low-dimensional factors. 
However, \emph{tensor network structure search (TN-SS)}---the task of selecting the optimal model-related hyperparameters such as TN ranks~\cite{Ye2019Tensor} and topology~\cite{li2020evolutionary}---can be a challenging problem due to its high-dimensional discrete nature and computational difficulty~\cite{hillar2013most}.

Until now, various approaches have been developed to solve TN-SS, employing different methods~\cite{hashemizadeh2020adaptive,li2020evolutionary,nie2021adaptive,chen2022one,li2022permutation,li2023alternating,zheng2023svdinstn,zeng2024bayesian}. 
Among these, sampling-based algorithms~\cite{hashemizadeh2020adaptive,li2020evolutionary,li2022permutation,li2023alternating} have demonstrated superior performance in addressing TN-SS for machine learning tasks.
 The core idea of these approaches is to sequentially draw samples in the search space according to their heuristic strategies, such as TNGA (\emph{evolutionary algorithm}, \citealp{li2020evolutionary}), GREEDY (\emph{greedy algorithm}, \citealp{hashemizadeh2020adaptive}), and TNLS and TnALE (\emph{local-search algorithms}, \citealp{li2022permutation,li2023alternating}), until optimal TN structures are reached. 
Additionally, these sampling-based approaches are more compatible with a variety of loss functions, model architectures, and optimizers in machine/deep learning.

However, achieving a good balance between exploration and exploitation in sampling-based heuristic approaches is notoriously difficult and seemingly endless.
This issue arises because different downstream tasks require different balance points, and various heuristic strategies have distinct preferences between exploration and exploitation.
For example, TNGA excels in exploration but performs poorly in exploitation, suffering from the curse of dimensionality. 
In contrast, TNLS and TnALE are more efficient than TNGA as they leverage the local smoothness prior of the search space.
However, as noted by \citet{li2023alternating}, they are more prone to getting stuck in local minima, particularly with real-world data.
This issue not only leads to poor performance of approaches but also results in a labor-intensive \emph{“research cycle”}, where human experts need to repeatedly develop new TN-SS algorithms to deal with different downstream tasks.


\emph{Is there a (semi-)automatic way to advance this cycle, reducing the need for intensive labor efforts?} If such a method exists, it would free human experts to tackle more challenging problems. More importantly, it would help us address the ultimate question: "How far can this cycle take us?"

To answer the questions, we propose a large language model (LLM)-driven automation framework, designed to automatically discover unknown TN-SS algorithms. 
Leveraging the remarkable full-domain knowledge of LLMs and their emergent understanding and reasoning capabilities,
the proposed framework, termed \underline{t}ensor-\underline{n}etwork-purposed \underline{GP}T-driven structure \underline{s}earch (tnGPS), takes the existing TN-SS algorithms (coded in Python) as inputs, and then mimic human experts' workflow for innovative research, to generate a batch of novel TN-SS algorithms expected to surpass the state-of-the-art (SOTA) algorithms.
It is worth noting that in tnGPS LLMs play a crucial role in ``innovation'': we construct a pipeline of prompts through which LLMs explore novel TN-SS algorithms from various perspectives, such as knowledge categorization (\texttt{KC}), knowledge recombination (\texttt{KR}), incremental innovation (\texttt{II}) and diversity injection (\texttt{DI}).
The generated new TN-SS algorithms are then evaluated in local computers with extensive downstream-task-specific numerical experiments.
The experimental results are subsequently used to update the prompts, guiding the LLMs to improve the algorithms in the next iteration.

We numerically evaluate the effectiveness of the discovered TN-SS algorithms by tnGPS on benchmarks, comparing them to the existing TN-SS algorithms. 
The experiment results demonstrate that tnGPS can discover novel TN-SS algorithms that not only outperform existing algorithms on in-domain data but also exhibit superior performance on out-of-domain data.
In addition, we also implement ablation experiments to analyze the contributions of various components of the tnGPS framework.
The main contributions of this paper are summarized as two-fold:
\begin{itemize}
[noitemsep,topsep=0pt,parsep=5pt,partopsep=0pt]
    \item We propose tnGPS, a large language model (LLM)-driven automation framework designed to automatically generate novel and effective TN-SS algorithms tailored to specific downstream tasks;
    \item {}Experimental results demonstrate that the algorithms discovered by tnGPS outperform existing TN-SS algorithms on benchmark data.

\end{itemize}

\subsection{Related Works}



\textbf{Tensor network structure search (TN-SS).}
The problem of TN-SS can be viewed as an extension of rank selection for tensor decomposition~\cite{rai2014scalable, zhao2015bayesian, yokota2016smooth}, which can be further traced back to studies of matrix factorization~\cite{6194350}.
Unlike rank selection, TN-SS aim to search for a richer set of model-related hyperparameters of a tensor network, including not only TN-ranks~\cite{cheng2020novel,mickelin2020algorithms,li2021heuristic,kodryan2023mars} but also network topology~\cite{hayashi2019exploring,hashemizadeh2020adaptive,li2020evolutionary, haberstich2023active} and permutation~\cite{chen2022one}.
From the technique perspective, various methods have been employed including Bayesian inference~\cite{zeng2024bayesian}, spectrum methods~\cite{chen2022one}, continuous optimization~\cite{zheng2023svdinstn} and discrete optimization~\cite{li2023alternating}, etc..
In this work, we follow the technical route of discrete optimization, particularly the sampling-based technique, due to its high precision and flexibility with a variety of loss functions.
Unlike the existing methods that solve TN-SS, the goal of this work is to develop a ``\emph{meta-method}'' that can create novel and effective TN-SS algorithms through the automatic discovery of algorithms.

\textbf{Automatic algorithm discovery (AAD).}
The studies on AAD can be traced back to the early 1900s~\cite{gratch1992composer,minton1993analytic}.
Building on this foundational works, more recent studies~\cite{khudabukhsh2016satenstein,meng2021automated,yi2022automated} developed various frameworks based on evolutionary programming and machine learning to discover efficient algorithms for solving computationally hard problems such as complex combinatorial optimization.
Parallel to these efforts, similar problems have been addressed in the field of AutoML~\cite{he2021automl}.
The difference is that the works~\cite{bello2017neural,real2020automl,wang2022efficient,chen2024symbolic} in AutoML dedicate efforts to neural architecture search and optimizer design.

\textbf{Large language models revolutionize AAD.}
LLMs have demonstrated remarkable performance in code programming~\cite{haluptzok2023language,liventsev2023fully,min2024beyond,hong2024metagpt,hemberg2024evolving,gur2023real}.
Since numerical algorithms are typically implemented on computers through programming languages like Python or C++, the capability of LLMs for code programming can be naturally extended to AAD~\cite{zelikman2023self,liu2023algorithm,pluhacek2023leveraging,liu2024example,ye2024reevo,romera2024mathematical}.
Note that conventional AAD methods discover new algorithms from a handcrafted, finite-dimensional algorithm space, while LLMs can explore new algorithms from an infinite-dimensional code space, which encompasses knowledge drawn from extensive training data across multiple fields. 
 The most closely related methods to ours are the works~\cite{liu2023algorithm,liu2024example} that utilize LLMs to evolve algorithms for solving the well-known traveling salesman problem.
Unlike their works, we target the TN-SS problem in this paper, establishing an AAD framework that leverages LLMs to mimic human experts.

\section{Preliminaries}
In this section, we first review the problem of tensor network structure search for self-constrained purposes.
Following this, we introduce a workflow that models how human experts typically innovate in research. This workflow will then be used in the next section to guide the design of the proposed framework.


\subsection{Tensor Network Structure Search (TN-SS)}
To provide intuitive simplicity, we review TN-SS through its application to tensor decomposition. 
Let $\mathcal{X}\in{}\mathbb{R}^{I_1\times{}I_2\times\cdots\times{}I_N}$ be a non-zero tensor of order $N$.
In this work, we consider solving TN-SS by minimizing the following objective function~\cite{li2020evolutionary}:
\begin{equation}
F(A):=  \phi(A)+ 
\lambda \cdot \min _{\{\mathcal{Z}_n\}_{n=1}^N} \underbrace{\|\mathcal{X}-tns(\{\mathcal{Z}_n\} ; A)\|_{F}^{2} /\|\mathcal{X}\|_F^2}_{\mbox{relative squared error (RSE)}}.
\label{eq:fitness}
\end{equation}
Here, $A\in\mathbb{A}\subseteq\mathbb{R}^{N\times{}N}$ represents the adjacency matrix that formulates the graphical diagram of a tensor network, \textit{i.e.}, the tensor network (TN) structure.
The feasible set $\mathbb{A}$ is determined by specific searching domains, such as TN-ranks, permutations~\cite{li2022permutation} and topology~\cite{li2020evolutionary}.
The set $\{\mathcal{Z}_n\}_{n=1}^N$ represents the collection of core tensors~\cite{cichocki2016tensor} of a tensor network $tns(\{\mathcal{Z}_n\} ; A)$ associated to the TN-structure $A$~\cite{Ye2019Tensor}.
The objective function~\eqref{eq:fitness} describes a linear combination of two terms with a tuning parameter $\lambda>0$, where the first term $\phi(A)$ models a TN's complexity (such as compression ratio and graph sparsity), and the second term calculates the minimization of relative squared error (RSE) in tensor decomposition, modeling the expressibility of a TN. 
In summary, within the context of tensor decomposition, TN-SS aims to find the optimal TN structures, modeled with $A$, that minimizes both the TN's complexity and the RSE.


\subsection{A Unified Paradigm for TN-SS Algorithms}
In this work, we consider minimizing the objective function~\eqref{eq:fitness} as a discrete optimization problem.
Most existing algorithms that solve \eqref{eq:fitness} follow a ``\emph{sampling-evaluation}'' paradigm~\cite{li2020evolutionary,li2022permutation,li2023alternating}.
Algo.~\ref{alg:TN-SS} provides its basic description, in which the key ingredient is the sampling operation as follows:
\begin{equation}
    p\leftarrow\textit{GenerateSamples}(C, P, F, i, m, \#Iter, L),\label{eq:GenerateSamples}
\end{equation}
where $p$ denotes a collection of adjacency matrices defined in~\eqref{eq:fitness}, the arrow $\leftarrow$ represents the value assignment, and  $C, P, F, i, m, \#Iter, L$ denotes arguments required in the function.
They contain important information such as the current optimal TN structures $C$, historical sampled structures and their corresponding evaluation scores $(P,F)$, as described in Algo.~\ref{alg:TN-SS}.
The operation~\eqref{eq:GenerateSamples} implies that in each iteration a batch of new adjacency matrices are generated, conditioned on the historical adjacency matrices and their corresponding evaluation scores in the searching process. 
Given adjacency matrices, the evaluation scores are calculated in the operation $Eval(\,p\,)$, by minimizing the objective~\eqref{eq:fitness} with respect to $\{\mathcal{Z}_n\}_{n=1}^N$.

Under the paradigm illustrated in Algo.~\ref{alg:TN-SS}, the existing TN-SS algorithms differ mainly from the customization on the $\textit{GenerateSamples}(\,\cdot\,)$ in~\eqref{eq:GenerateSamples}.
For example, TNGA~\cite{li2020evolutionary} specifies \eqref{eq:GenerateSamples} with evolutionary operators; GREEDY~\cite{hashemizadeh2020adaptive}, TNLS~\cite{li2022permutation} and TnALE~\cite{li2023alternating} leverage incremental, random or alternating sampling in neighborhood to generate new samples.
In this work, we aim to leverage LLMs to discover new ideas for designing the function $\textit{GenerateSamples}(\,\cdot\,)$, with the goal  of improving the performance of solving TN-SS.

\input{modules/algo1}

\subsection{How Human Experts do Innovation?}
\input{modules/fig_workflow}
The intuition of this work is to harness LLMs to mimic human experts to realize the goal of discovering novel TN-SS algorithms.
In doing so, we formalize below a basic but complete workflow, shown in Figure~\ref{fig:workflow}, to describe how human experts do innovative research.

In academic research, human experts typically begin with preliminary studies, such as gathering information through literature reviews and paper retrieval.
The gathered information is then compiled into an \emph{idea pool}, which includes numerous existing ideas related to the targeted problem, such as solving TN-SS.
Each idea is accompanied by various descriptors, including an evaluation score and contextual information that describes the idea’s background, intuition, and other relevant details.
Subsequently, in the \emph{knowledge categorization} \texttt{(KC)} phase, the ideas are refined into \emph{knowledge} clusters. 
Each cluster consists of different ideas that follow a similar hypothesis or principle.
In \emph{Idea Dropout} \texttt{(ID)}, this phase models the behavior of human experts by filtering out certain ideas from the pool, allowing them to focus only on the most interesting ones for deeper study.
The metrics used in \texttt{ID} are typically multivariate, encompassing factors such as personal interests, performance, trends, or even randomness due to the large scale of the idea pool.

In this work, innovation is considered as an operation that generates new ideas from existing ones.
As shown in Figure~\ref{fig:workflow}, we model innovation as a two-stage process consisting of
\emph{knowledge recombination} \texttt{(KR)} and \emph{incremental innovatio} \texttt{(II)}
These phases are key ingredients to create values in not only research but also in strategy and entrepreneurship~\cite{rubin2018creating,xiao2022knowledge}.
More specifically speaking, \texttt{KR} refers to the process of generating new ideas by merging, integrating, or reconfiguring existing ideas.
\texttt{II} involves gradual improvements or minor modifications to existing ideas.
Additionally, Figure~\ref{fig:workflow} formalizes \emph{diversity injection} \texttt{(DI)}, which is forced to generate new ideas that are ``orthogonal'' to the existing ones.
\texttt{DI} is commonly observed in real-world scenarios, such as brainstorming sessions in team meetings or critical comments from non-experts.
Significant innovation is ultimately expected to emerge by recursively invoking \texttt{KR}, \texttt{II}, and \texttt{DI}  within the workflow.



\section{tnGPS: a LLM-Driven Framework for TN-SS Algorithm discovery}
In this section, we introduce tnGPS, an LLM-driven framework for discovering TN-SS algorithms. The introduction primarily focuses on the technical aspects, illustrating how tnGPS is designed with a pipeline of prompts to harness LLMs in order to discover novel TN-SS algorithms, mimicking the innovation process of human experts.

\textbf{Global architecture of tnGPS.}
We conceptualize tnGPS as a system where the inputs are known TN-SS algorithms encoded in Python, and the outputs are newly discovered TN-SS algorithms. Inspired by the way human experts conduct innovative research, tnGPS is designed with a global pipeline similar to the one depicted in Figure~\ref{fig:workflow}. In this design, the LLM acts as an agent to perform the functions of each phase shown in Figure~\ref{fig:workflow}, replacing human experts.
Below, we introduce the specific implementation of each block in detail.


\textbf{Idea Pool}, also referred to as \emph{“the pool”} for brevity throughout the paper, is defined as a set of algorithms described as (\texttt{algorithm, score}). Here, \texttt{algorithm} contains various implementations of the function $\textit{GenerateSamples}(\,\cdot\,)$ as defined in Eq.~\eqref{eq:GenerateSamples} of a TN-SS algorithm using Python, and \texttt{score} is a scalar value indicating the algorithm’s performance in the TN-SS problem.
To help LLMs understand TN-SS algorithms effectively, all \texttt{algorithm}s in the pool are standardized with a unified interface, as depicted in Figure~\ref{fig:model_description}. Additionally, the text in Figure~\ref{fig:model_description} will serve as the “pilot” for constructing the prompt, as discussed at the end of this section.

\textbf{Knowledge categorisation} (\texttt{KC}).
In this phase, we create clusters of algorithms using LLMs. 
Each cluster contains similar \texttt{algorithms}, representing a distinct piece of knowledge on how to solve TN-SS.
In doing so, we first simplify and initialize the clusters with each algorithm in the pool. 
Once a new algorithm (e.g. one generated by tnGPS) is coming, we prompt the LLM to evaluate the methodological similarity between the new algorithm and the cluster centroids, which are the algorithms with the best scores in each cluster.
The new algorithm is then assigned an index to declare its cluster membership.
The key prompt used in this phase is illustrated in Figure~\ref{fig:cluster}.

\input{modules/functionDescription}
\input{modules/cluster}




\textbf{Idea dropout} (\texttt{ID}).
The dropout is conducted by randomly selecting algorithms from the pool using a roulette selection mechanism.
Consider $N$ algorithms, indexed from $1$ to $N$ based on their scores, ranked from highest to lowest.
Inspired by the previous work~\cite{li2020evolutionary}, the roulette selection performs a sequential sampling without replacement
The probability of selecting the $k$th algorithm is given by
\begin{equation}
   Pr(k)=\max \left\{0.01, \ln \left(\frac{\alpha}{ eps + k}\right)\right\}, \label{eq:sampling}
\end{equation}
where $eps$ is a very small positive number to ensure numerical stability, $\ln(\,\cdot\,)$ denotes the natural logarithm function, and $\alpha>0$ is a hyperparameter that controls the selection preference.
A smaller value of $\alpha$ increases the likelihood of selecting algorithms with higher scores.



To select a preferred algorithm in \texttt{ID}, we need to implement the roulette selection twice.
First, we perform selection at the knowledge level.
This involves selecting the centroids of clusters to determine which clusters will be considered.
Second, within the selected clusters, we perform selection at the algorithm level to choose the preferred algorithm from each cluster.
This “bi-level” process is repeated until the desired number of algorithms has been selected.

\textbf{Knowledge recombination} (\texttt{KR}).
Guided by the workflow shown in Figure~\ref{fig:workflow}, \texttt{KR} is a fundamental phase to generate novel TN-SS algorithms in tnGPS.
In this phase, the input consists of $N\geq{}0$ pairs of (\texttt{algorithm, score}).
The output is $M$ new algorithms generated by the LLM.
The key prompt used in \texttt{KR} is illustrated in Figure~\ref{fig:KR}.
The design of this prompt aims to enable the LLM to understand the factors contributing to the superior performance of certain algorithms while avoiding the meaningless stacking of Python code.

\input{modules/KR}



\textbf{Incremental innovation} (\texttt{II}) follows \texttt{KR} as another block for creating innovation in tnGPS.
Unlike \texttt{KR} which recombines algorithms,  \texttt{II} aims at mortifying algorithms individually and slightly.
Figure~\ref{fig:II} gives the key prompt used in this phase.
In our experience with GPT-3.5/4, we found that the LLM tends to improve code aspects such as efficiency, readability, and parallelism, which are not the focus of this phase. To prevent this, we include specific instructions (highlighted in orange in Figure~\ref{fig:II}) to guide the LLM away from these unintended improvements and focus on the targeted modifications.

\input{modules/II}


\textbf{Diversity injection}~(\texttt{DI}).
We implement \texttt{DI} by leveraging the LLM to create new clusters in the algorithm pool. 
In doing so, we design the prompt as Figure~\ref{fig:DI}, instructing the LLM to generate TN-SS algorithms that are methodologically distinct from the centroids of existing clusters.
Following this, the newly created algorithms from \texttt{DI} serve as centroids for new clusters.
The newly created algorithms from \texttt{DI} serve as centroids for new clusters.
Since relying solely on existing knowledge can lead to “path dependence”—where new ideas are heavily influenced by past knowledge—we encourage the LLM to explore new ideas in \texttt{DI} without considering performance metrics (e.g., scores). This approach enriches the diversity of the algorithm pool.
\input{modules/DI}


\textbf{Format restriction and prompt architecture.}
In addition to the goal-oriented prompts mentioned earlier, we need to enforce specific format restrictions on the LLM’s output. For instance, this includes omitting unnecessary analyses and specifying the desired format for code outputs. Furthermore, instructions related to code writing must be included to ensure that there are no compilation failures during evaluation.
To meet these requirements, we designed the prompt shown in Figure~\ref{fig:FR}. It’s important to note that the prompt in Figure~\ref{fig:FR} was developed through a trial-and-error process, making it dependent on the specific LLM used and our code-writing conventions. As a result, manual adjustments may be necessary when using different LLM models.

\input{modules/FR}

Finally, the complete prompts used in \texttt{KC, KR, II} and \texttt{DI} are constructed following the same architecture as shown in Figure~\ref{fig:frame}.
It consists of three parts: interface description (Figure~\ref{fig:model_description}), those goal-oriented prompts following a list of algorithms and scores (e.g., Figures~\ref{fig:cluster},~\ref{fig:KR},~\ref{fig:II},~\ref{fig:DI}), and format restriction (Figure~\ref{fig:FR}).
These three parts are concatenated together to provide comprehensive instructions to the LLM.



\input{modules/frame}

\textbf{Experiments (evaluation) with sandbox.}
All newly generated TN-SS algorithms are evaluated on local (super)computers using task-specific training data. However, due to the probabilistic nature of LLMs, there is a possibility of receiving unexpected outputs, such as unrunable code, unnecessary comments, or error messages from the LLM platforms.
To address this issue, we construct a sandbox environment to run the code in a relatively isolated setting using small-scale training data before conducting high-cost formal numerical experiments. If program errors, abnormal resource consumption, or unexpected output formats occur, the sandbox will immediately terminate the process and remove the problematic code from the job queue awaiting implementation.

\section{Experimental Results}
In this section, we use several benchmarks to demonstrate that tnGPS can discover new algorithms that outperform  SOTA methods for TN-SS.
Additionally, we conduct ablation experiments to evaluate the impact of each component within tnGPS on the discovery performance.


\subsection{Natural Images Compression}
In this experiment, we aim to use TN-SS algorithms to search for optimal topology and ranks for a tensor network (TN) to represent natural images with fewer parameters.


\textbf{Data preparation.} 
We randomly select $14$ images from the BSD500 dataset \cite{arbelaez2010contour}.
These images are converted to grayscale and resized to $256 \times 256$ pixels.
Each image is then reshaped into an order-$8$ tensor by the default Python reshaping function.
The $14$ pre-processed images are split into two sets: $4$ images for training and $10$ images for testing. 

\textbf{Settings of tnGPS.}
We use three existing TN-SS algorithms as inputs: TNGA~\cite{li2020evolutionary}, GREEDY~\cite{hashemizadeh2020adaptive} and TNLS~\cite{li2022permutation}.
The evaluation phase in tnGPS calculates Eq.~\eqref{eq:fitness} for each generated algorithm, averaging the results over the images in the training set.
In Eq.~\eqref{eq:fitness}, we set $\lambda=5$ and use the same compression ratio function $\phi$ as in previous work~\cite{li2020evolutionary}.
The hyperparameters required in tnGPS are listed in Table~\ref{tab:para}.
For this experiment, we set $m=2$, $n=1$, $\alpha_{1}=\alpha_{2}=100$, $c=5$ and $\#Iter=30$.
Additionally, we select \texttt{gpt-4-1106-preview} as the LLM model, applying a temperature of $0.7$ uniformly across all prompts. 
After implementation, we select the top-three algorithms from the pool (excluding the input algorithms), termed Ho-1\footnote{The name ``Ho'' is shorthand for \emph{''homunculus''}, representing that these algorithms are created through some ``unusual'' means.}, Ho-2, and Ho-3, as the outputs of tnGPS.
The three algorithms will be evaluated and compared with the existing TN-SS algorithms.

\input{modules/para}


\begin{figure*}[t!]
    \centering
\includegraphics[width=2.082\columnwidth]{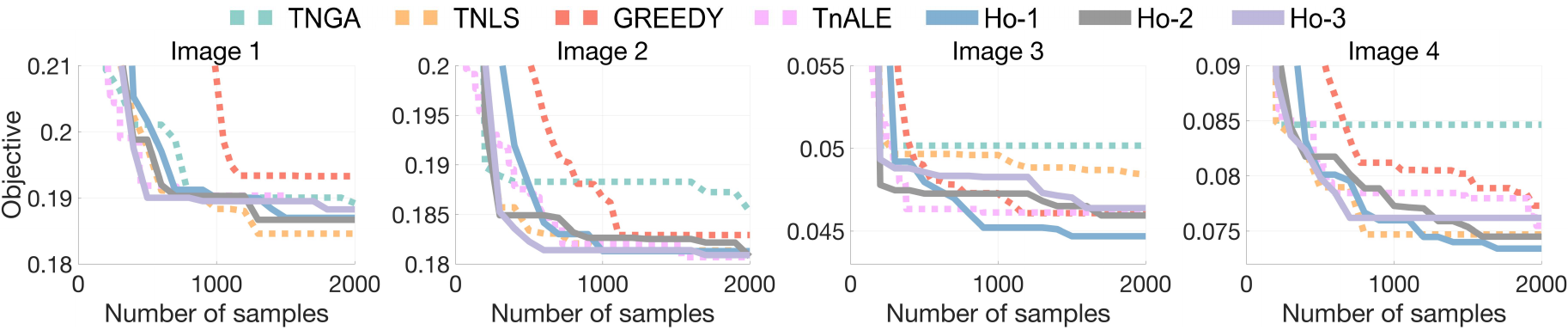}
    \vspace{-0.6cm}
    \caption{Objective $vs.$ number of sample curves of different algorithms on four training images.
    }
    \label{fig:nature_train}
    \vspace{-0.0cm}
\end{figure*}
\begin{figure*}[t!]
    \centering
\includegraphics[width=2.082\columnwidth]{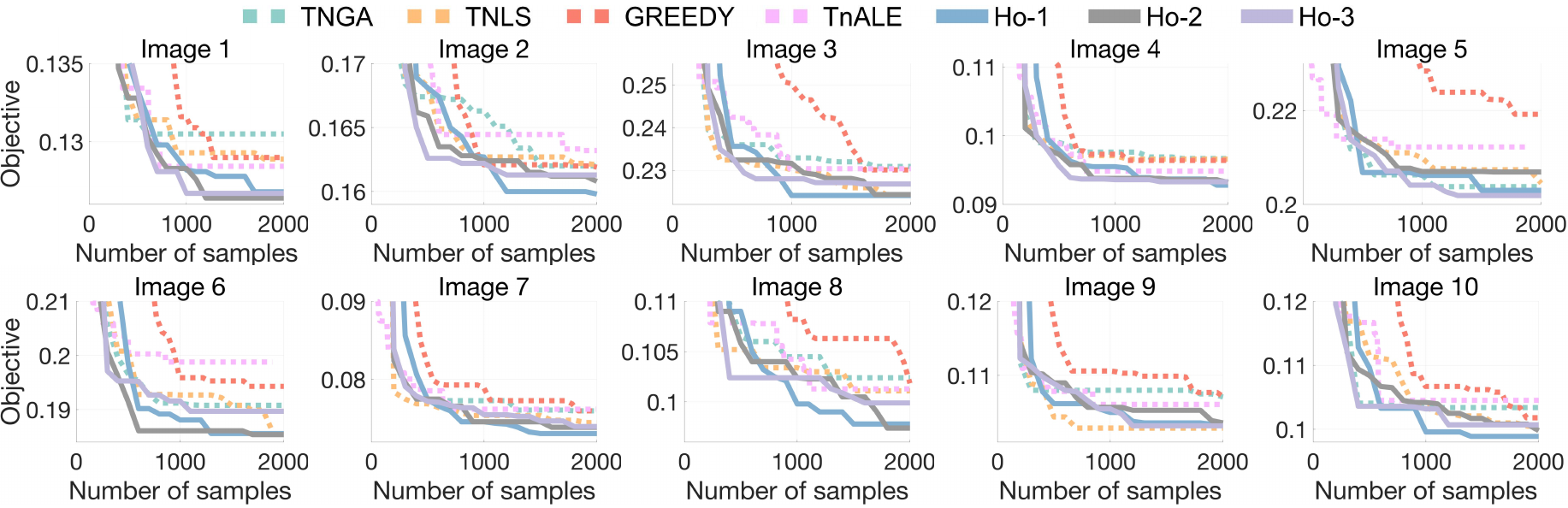}
    \vspace{-0.6cm}
    \caption{Objective $vs.$ number of sample curves of different algorithms on ten testing images.
    }
    \label{fig:nature_test}
    \vspace{-0.4cm}
\end{figure*}

\begin{table*}[t]
	\centering
	\caption{The averaged compression ratio (in $\log$ form, calculated by dividing the number of parameters of the image by the number of parameters of the TN) and the RSE for the structures obtained by the algorithms.
	}
	\begin{threeparttable}\small\label{tab:natureimage}
		\setlength{\tabcolsep}{2.5mm}{   	
			\begin{tabular}{ccccc|ccc}
				\toprule
	\multirow{3}[0]{*}{\textbf{Images}}
			&\multicolumn{7}{c}{\textbf{  Log compression ratio$\uparrow$ $+$ RSE$\downarrow$} -- \emph{CR(RSE)}}\\
				\cmidrule{2-8} 
    &\textbf{TNGA} &  \textbf{TNLS} & \textbf{GREEDY}& \textbf{TnALE} & \textbf{Ho-1} & \textbf{Ho-2}&\textbf{Ho-3}
\\
				\midrule
				\textbf{Train} &1.478~(0.128)&1.419~(0.119)&1.451~(0.123)&1.439~(0.122)&1.436~(0.120)&1.496~(0.124)&1.446~(0.121)\\
				\textbf{Test} &1.332~(0.132)&1.335~(0.131)&1.331~(0.132)&1.328 ~(0.132)&1.329~(0.129)&\textbf{1.352~(0.130)}&1.322~(0.129)\\
    
				\bottomrule
			\end{tabular}
		}
	\end{threeparttable}
\end{table*}

					

						   

\begin{table*}[t]
\caption{Number of parameters ($\times{}1000$) for TGP model compression. The values in [square brackets] give the number of samples used to find the structure for the first time. 
The symbol ``-'' means the algorithm fails to achieve the experiment's configuration.}
\vspace{0.3cm}
\centering\small
	\begin{threeparttable}\centering
		\setlength{\tabcolsep}{0.8mm}{   	
			\begin{tabular}{cc|ccccc|ccc}
				\toprule
					
			&\textbf{Initialization}&\textbf{Baseline} & \textbf{TNGA} & \textbf{TNLS}& \textbf{GREEDY}&\textbf{TnALE} & \textbf{Ho-1} & \textbf{Ho-2} & \textbf{Ho-3}  \\
				\midrule			
			CCPP&\multirow{2}{*}{\texttt{random}}&2.64&2.36~[1900]&2.50~[1900]&2.60~[850]&2.36~[588]&2.60~[1900]&\textbf{2.24}~[1600]&2.74~[1300]\\
            MG&&3.36&12.69~[8400]&17.25~[7600]&-&-&6.81~[9200]&\textbf{3.01}~[10000]&27.74~[8400]\\

      \midrule
      CCPP&\multirow{2}{*}{\texttt{TT}}&2.64&	2.24~[500]&2.24~[200]&2.24~[21]&2.24~[\textbf{18}]&2.24~[400]&2.24~[500]&2.24~[300]\\
MG&&3.36&3.36~[100]&3.01~[500]&3.01~[64]&3.01~[\textbf{42}]&3.01~[500]&3.01~[1200]&3.01~[2400]\\

				\bottomrule
						   
			\end{tabular}
 		}
	\end{threeparttable}\label{tab:GP}
	\vskip -0.1in
	\end{table*}

					
						   

\begin{table}[t]
\caption{Value of objective function~\eqref{eq:fitness} of the best algorithm discovered by tnGPS and its variants. In the table, ``baseline'' refers to the best result obtained from TNGA, TNLS, GREEDY, and TnALE, ``tnGPS'' refers to the proposed model, and ``\texttt{KR, II, DI}'' are tnGPS's variants, whose corresponding components are ablated.}
\vspace{0.3cm}
\centering\small
	\begin{threeparttable}\centering
		\setlength{\tabcolsep}{1.5mm}{   	
			\begin{tabular}{cccccc}
				\toprule				
			&\textbf{baseline} & \textbf{tnGPS} & \texttt{KR}&\texttt{II}&\texttt{DI} \\
				\midrule			
	\textbf{Objective}&0.1558&\textbf{0.1102 }&0.1308 &0.1273 & 0.1239\\
				\bottomrule	   
			\end{tabular}
 		}
	\end{threeparttable}\label{tab:ablation_tnGPS}
	\vskip -0.1in
\end{table}

\begin{table}[t]
\caption{Value of objective function~\eqref{eq:fitness} of the best algorithm discovered by tnGPS and its variants. In the table, ``baseline'' refers to the best result obtained from TNGA, TNLS, GREEDY, and TnALE, ``GPT-4, GPT-3.5, Claude-1, Claude-2'' refers to tnGPS using various LLMs, and ``Incomplete descriptions'' is the variant of tnGPS, whose interface description (Figure~\ref{fig:model_description}) is partially removed.}
\vspace{0.3cm}
\centering\small
	\begin{threeparttable}\centering
		\setlength{\tabcolsep}{1.2mm}{   	
			\begin{tabular}{cccccc}
				\toprule				
			\textbf{baseline} & \textbf{GPT-4} & \textbf{GPT-3.5}&\textbf{Claude-1} & \textbf{Claude-2}&\textbf{\makecell{Incomplete \\ descriptions}} \\
				\midrule			
			0.1847&\textbf{0.1813}&0.1842&0.1840&0.1834&0.1819\\
				\bottomrule	   
			\end{tabular}
 		}
	\end{threeparttable}\label{tab:ablation_LLMs_ID}
	\vskip -0.1in
\end{table}

\textbf{Implementation details.}
In the experiment, we implement four additional sampling-based TN-SS algorithms including TNGA~\cite{li2020evolutionary}, TNLS~\cite{li2022permutation}, GREEDY~\cite{hashemizadeh2020adaptive} and TnALE~\cite{li2023alternating}. 
Since the vanilla TNGA and TNLS are designed to search only for the topology or permutation of a TN, we extend them to fit the settings of this experiment.
Specifically, we extend TNGA by relaxing its binary constraint for encoding the topology with integers as done in the previous works~\cite{li2022permutation,li2023alternating}.
For TNLS, we fix the permutation and set the \emph{template} used in the algorithm to be a complete graph, enabling simultaneous search for TN topology and ranks.
In GREEDY, we modify its objective function from RSE to the function in \eqref{eq:fitness}, and further allow the algorithm to both increase and decrease the ranks during the search.


The parameter settings of the algorithms are as follows:
in TNGA, we set $\alpha = 100$, $\beta = 5$, the elimination rate to $10\%$, and the mutation probability to $25\%$;
in TNLS, we set $c_{1}=0.99$;
and in TnALE, we set $L_{0}=0, L=15, r_{2}=1$, and $D=1$.
For all methods (including those generated by tnGPS), we set the upper limit for rank search to $4$, the number of iterations in searching to $20$, and the number of samples\footnote{In TnALE, the number of samples in each iteration is determined by other hyperparameters.} in each iteration to $100$.
For TNGA and the algorithms generated by tnGPS, we initialize them with same TN structures, which have TN ranks close to one but with a $15\%$ probability of changing each rank from $1$ to $2$. 
We then select the best TN structure from the TNGA initialization to initialize GREEDY, TNLS, and TnALE.



\textbf{Results.}
Figures~\ref{fig:nature_train} and \ref{fig:nature_test} show how the value of the objective function~\eqref{eq:sampling} changes with increasing the number of samples for different TN-SS algorithms.
Table~\ref{tab:natureimage} shows the averaged performance metrics, including compression ratio and RSE, of different algorithms on both the training and test sets.
As shown in Table~\ref{tab:natureimage}, the three algorithms generated by tnGPS achieve comparable performance on the training set, while `Ho-2'' outperforms other algorithms on average in the test set.
Figure~\ref{fig:nature_test} visually demonstrates that the curves associated with the three algorithms generated by tnGPS tend to reach smaller values of the objective function compared to other methods.



\textbf{New insights gained from the generated algorithms.}
The codes of Ho-1,2,3 are provided in Appendix \ref{appendix:code}.
These codes reveal several new insights on how to improve the effectiveness and efficiency of solving TN-SS problems.
First, the new algorithms are no longer “Markovian”. Unlike previous methods that considered only the samples from the last iteration, the new algorithms incorporate information from all historical samples.
Second, the annealing trick is used inversely. For instance, in Ho-1, the algorithm employs a mutation operation similar to TNGA, with the mutation rate updated as in TNLS. However, contrary to the traditional annealing trick used in TNLS, the mutation rate in Ho-1 increases progressively to enhance exploration.
Third, the new algorithms adopt various novel exploitation strategies inspired by TNLS. Notably, Ho-2 introduces an innovative Gaussian perturbation mutation strategy, which, to the best of our knowledge, is unprecedented in the existing genetic algorithms literature.

\subsection{Model Compression for Gaussian Process: an Out-of-Domain Experiment} 
In this experiment, we evaluate whether the algorithms generated by tnGPS maintain their effectiveness for the out-of-domain tasks, \textit{i.e.}, tasks not considered in the algorithm discovery process.
In doing so, we follow previous TN-SS works~\cite{li2022permutation,li2023alternating} and consider the compression of a regression model using tensorial Gaussian process (TGP) \cite{izmailov2018scalable}.


\textbf{Experiment Setup.} 
We apply TN-SS to the high-order variational mean tensor of TGP to compress the model while preserving the prediction accuracy.
Two datasets are used: CCPP \cite{tufekci2014prediction} and MG \cite{flake2002efficient}.
In these datasets, the corresponding variational means are tensors of order 4 with a mode dimension of 12, and order 6 with a mode dimension of 8, respectively.
For the experiment, the upper bound for rank search is set to $10$. The Adam~\cite{kingma2014adam} learning rate is set to $0.001$, and the core tensors of TN are initialized with a Gaussian distribution with a variance of $0.01$.
Due to the distinct scales of the two datasets, the hyperparameter $\lambda$ in Eq.~\eqref{eq:fitness} is set to $10^{5}$ for CCPP and $10^{7}$ for MG, respectively.

We consider two types of initialization in the experiment: \texttt{random} and \texttt{TT}.
In \texttt{random}, the initial TN structures are selected randomly following the same method used in the preceding experiment.
In \texttt{TT}, we first initialize TN structures using the method in \texttt{random}, but one of the initialized structures is then replaced by tensor train~\cite{oseledets2011tensor}, which is the model used in the baseline~\cite{izmailov2018scalable}.
If only one structure requires initialization, such as in TNLS and TnALE, the baseline tensor train is used directly.
In the settings of MG $+$ \texttt{random}, we set the number of iterations in searching to $50$, and the number of samples in each iteration to $200$.
In other settings, we set the number of iterations in searching to $30$, and the number of samples in each iteration to $100$.


\textbf{Results.} 
The experimental results are presented in Table~\ref{tab:GP}, which shows the number of parameters (in thousands) required by tensor networks to represent the model.
Additionally, the number of samples used by the algorithms to find the structure for the first time is indicated in square brackets.
We can see that, in the \texttt{TT} initialization, all algorithms find better structures than the baseline, with GREEDY and TnALE requiring fewer samples than the other algorithms.
However, in the \texttt{random} initialization, only Ho-2 finds structures as good as those in \texttt{TT}. 
Although GREEDY and TnALE still converge quickly, the solutions they find are worse than Ho-2's.
Notably, in the setting of MG+\texttt{random}, only Ho-2 finds a structure better than the baseline.
These results suggest that tnGPS can generate new TN-SS algorithms that consistently outperform the existing SOTA methods in both in-domain and out-of-domain tasks. 


\subsection{Ablations} 
Next, we conduct ablation studies to evaluate the impact of various components of tnGPS on its performance.
These components include those LLM-driven phases in the pipeline, the interface description, and the selection of LLM models.
By systematically removing or modifying these components, we aim to understand their individual contributions to the overall effectiveness of tnGPS.

\textbf{LLM-driven phases in tnGPS.}
Here, we individually ablate the phases including knowledge recombination (\texttt{KR}), incremental innovation (\texttt{II}), and diversity injection (\texttt{DI}). 
We use the variational mean tensor from the MG dataset in the model compression experiment as the training data for tnGPS to assess the impact of each phase on the overall performance.

In this part, the hyper-parameter settings for the tnGPS are the same as in the image compression experiment. The other TN-SS algorithms maintain the same settings as in the model compression experiment, except that we set the number of samples per iteration to 50 and run the algorithms for 10 iterations for simplicity.

We present the experimental results for the tnGPS components in Table \ref{tab:ablation_tnGPS}. 
As observed, the complete tnGPS model achieves the best results, surpassing existing methods. 
However, the model's performance declines when components like \texttt{KR}, \texttt{II} and \texttt{DI} are removed, underscoring the importance of integrating, enhancing, and injecting ideas. Additionally, the ablation results for \texttt{KR}, \texttt{II} and \texttt{DI} still surpass existing methods, suggesting the potential benefits of these individual components in discovering better algorithms.

\textbf{Selection of LLM models and interface description.} 
 We next examine how the selection of LLM models affects tnGPS's performance. 
We compare GPT-4 (\texttt{gpt-4-1106-preview}), GPT-3.5 (\texttt{gpt-3.5-turbo-16k-0613}), Claude-1 (\texttt{claude-instant-1}), and Claude-2 (\texttt{claude-2}). 
 Additionally, we analyze tnGPS's sensitivity to interface description (shown in Figure~\ref{fig:model_description}) by randomly removing 8 from the total 12 comments (see Figure~\ref{fig:comments_removal} in Appendix). 
For simplicity, we select image 2 from the training set in the image compression experiment as the training data and use the same hyper-parameter settings as in the image compression experiment, except for adjusting the iteration and sample settings for the TN-SS algorithm as done in the LLM-driven phases ablation experiment.

As concluded from the results in Table \ref{tab:ablation_LLMs_ID}, more powerful LLMs like GPT-4 lead to better tnGPS performance.
Moreover, by further analyzing the algorithms generated by GPT-3.5, we found that it tends to make trivial modifications of algorithms, such as the naive stacking of different methods. Additionally, the incomplete interface description does not significantly affect the performance of tnGPS. 
We conjecture that the algorithms themselves provide sufficient information to let LLMs understand the function to be generated. 
However, during the experiment, we observed that the LLM sometimes misunderstood the meaning of the objective function values, for example, mistakenly thinking that larger values implied better performance.


\section*{Concluding Remarks}
Our experiential results confirm the positive impact of LLMs on solving TN-SS. 
Specifically, the proposed framework, tnGPS, can leverage insights gained from the existing algorithms and the embedded knowledge in LLMs to automatically discover novel TN-SS algorithms that achieve a better balance between exploration and exploitation. 
The benchmarks of image compression and model compression consistently demonstrate the superiority of the discovered algorithms in finding better TN structures.

\textbf{Limitation.} A primary limitation of our method is that the final discovered algorithms' performance is subject to variation due to changes in LLMs. Nonetheless, we anticipate a positive correlation between the performance of designed algorithms by our method and the capabilities of LLMs.

\section*{Impact Statement}
This paper explores the potential of using LLMs to automatically discover improved algorithms without human intervention, a step that could significantly boost the efficiency of algorithm development and augment production processes. However, it also raises concerns about LLMs generating unreliable results due to the hallucination issue, which affects their reliability and practicality. To address this, our study emphasizes the significance of not only evaluating the outputs produced by LLMs but also constructing precise prompts to guide them effectively. These approaches help minimize the risk of unreliable results, ensuring the practicality and reliability of LLMs for practitioners. Moreover, the increasing capabilities of LLMs may also lead to the question of whether machines may eventually replace human expertise entirely. 

\section*{Acknowledgements}
We appreciate the anonymous (meta-)reviewers for their helpful comments and are grateful to Yumeng Ma for her invaluable advice in formulating the method.
Junhua completed this work during an internship at RIKEN-AIP under the supervision of Chao and Qibin.
The work was partially supported by National Natural Science Foundation of China (Grant No. 62073087 and 62071132), and JSPS Kakenhi (Grant No. 24K03005 and JP23K28109).
Additionally, Chao was also supported by RIKEN Incentive Research Project.




\nocite{langley00}

\bibliography{example_paper}
\bibliographystyle{icml2024}

\newpage
\appendix
\onecolumn
\input{sections/99_appendix}

\end{document}

%% file: sections/00_abstract.tex
\begin{abstract}

Tensor networks are efficient for extremely high-dimensional representation, but their model selection, known as tensor network structure search (TN-SS), is a challenging problem.
Although several works have targeted TN-SS, most existing algorithms are manually crafted heuristics with poor performance, suffering from the curse of dimensionality and local convergence.
In this work, we jump out of the box, studying how to \emph{harness large language models (LLMs) to automatically discover new TN-SS algorithms, replacing the involvement of human experts.}
By observing how human experts innovate in research, we model their common workflow and propose an automatic algorithm discovery framework called tnGPS.
The proposed framework is an elaborate prompting pipeline that instruct LLMs to generate new TN-SS algorithms through iterative refinement and enhancement.
The experimental results demonstrate that the algorithms discovered by tnGPS exhibit superior performance in benchmarks compared to the current state-of-the-art methods.
Our code is available at \url{https://github.com/ChaoLiAtRIKEN/tngps}.

\end{abstract}

%% file: modules/algo1.tex
\begin{algorithm}[tb]
\caption{The ``Sampling-Evaluation'' Paradigm}
\label{alg:TN-SS}
\begin{algorithmic}[1] 
\State \textbf{Initialize:}
\State $m$ \Comment{Number of samples}
\State $L$ \Comment{Set of hyperparameters}
\State $\#Iter$ \Comment{Maximum number of iterations}
\State $P \gets []$ \Comment{Historical TN structures, initially empty}
\State $F \gets []$ \Comment{Historical evaluation scores, initially empty}
\State $C \gets []$ \Comment{The best TN structures, initially empty}

\State
\For{$i = 1$ \textbf{to} $\#Iter$}
    \State $p \gets \textit{GenerateSamples}(C, P, F, i, m, \#Iter, L)$
    \State $F \gets F \cup \textit{Eval}(p)$ \Comment{Evaluate new samples}
    \State $P \gets P \cup p$ \Comment{Update historical samples}
    \State Update $C$ if necessary \Comment{Update the best structure}
\EndFor
\State
\State \textbf{Output:} Set $C$ containing the  best TN structures.
\end{algorithmic}
\end{algorithm}

%% file: modules/fig_workflow.tex
\begin{figure}
    \centering
\includegraphics[width=1\columnwidth]{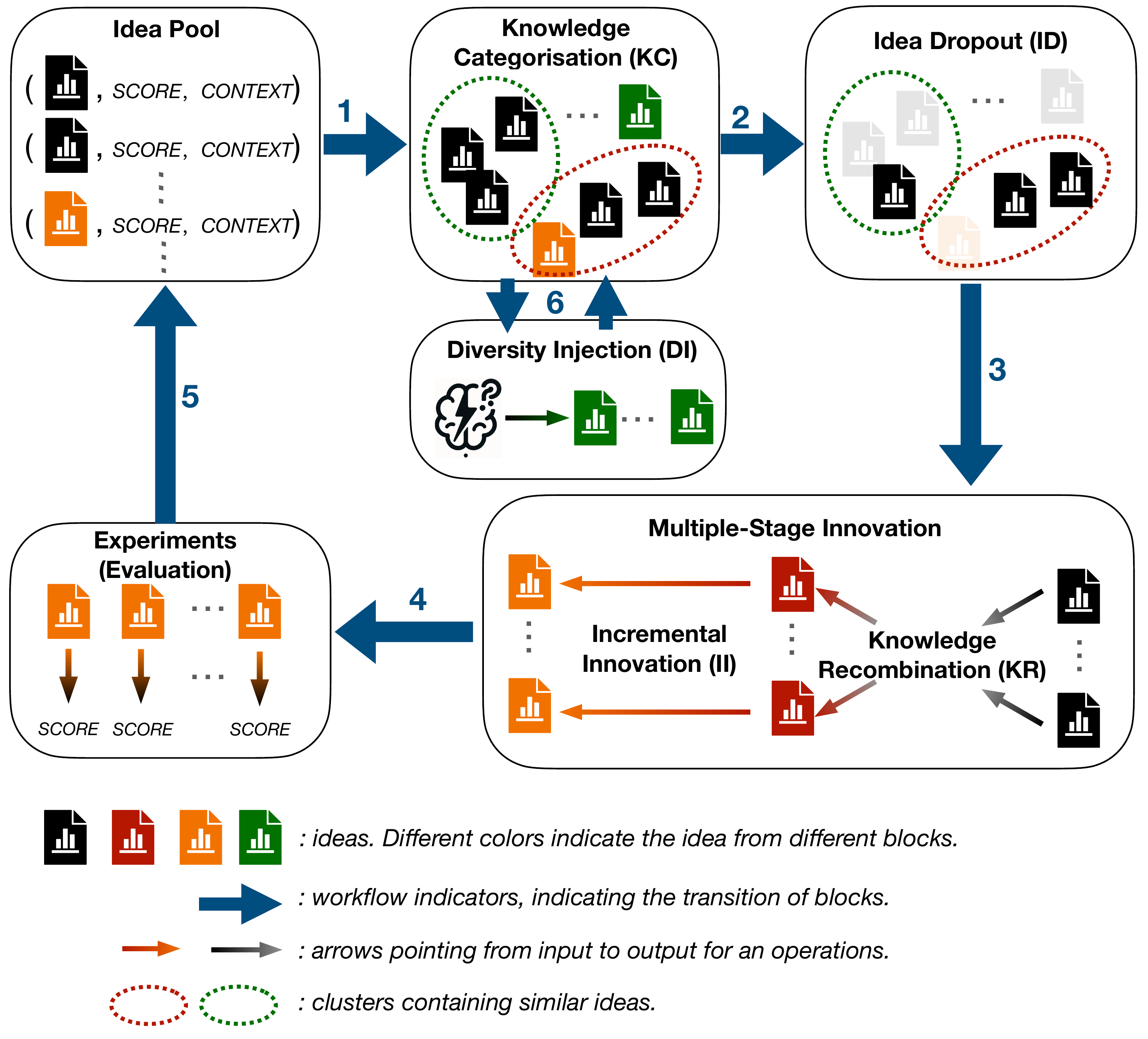}
    \vspace{-0.4cm}
    \caption{
A basic workflow to illustrate how human experts do innovation in research.
    }
    \label{fig:workflow}
    \vspace{-0.4cm}
\end{figure}

%% file: modules/functionDescription.tex
\begin{figure}
    \centering
\includegraphics[width=1\columnwidth]{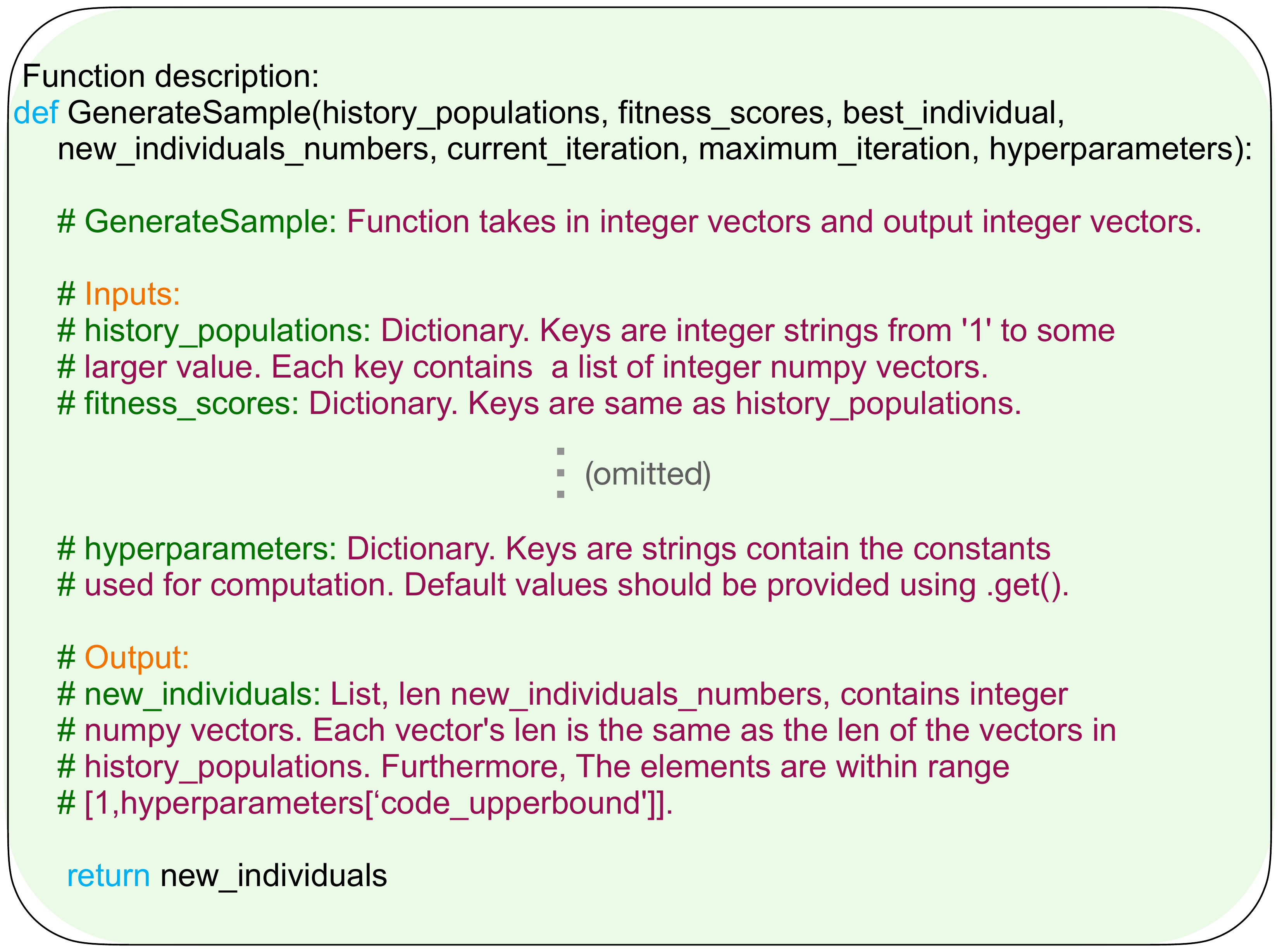}
    \vspace{-0.4cm}
    \caption{
The prompt used for interface description.
    }
    \label{fig:model_description}
\end{figure}

%% file: modules/cluster.tex
\begin{figure}[h]
    \centering
    \includegraphics[width=1\columnwidth]{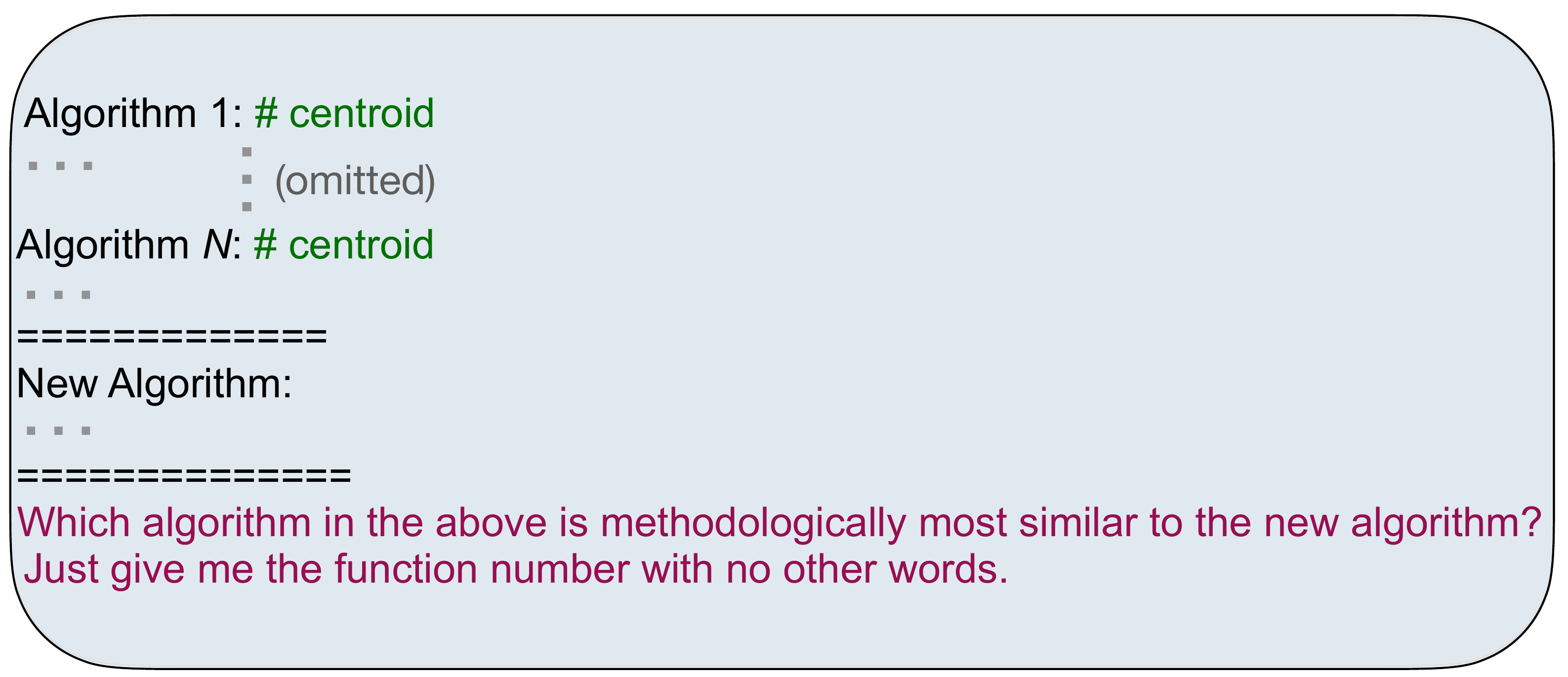}
    \vspace{-0.4cm}
    \caption{
The key prompt used in knowledge categorisation (\texttt{KC}).
The \emph{\color{purple}purple} sentence specifies the goal of the prompt.
    }
    \label{fig:cluster}
    \vspace{-0.4cm}
\end{figure}

%% file: modules/KR.tex
\begin{figure}[h]
    \centering
    \includegraphics[width=1\columnwidth]{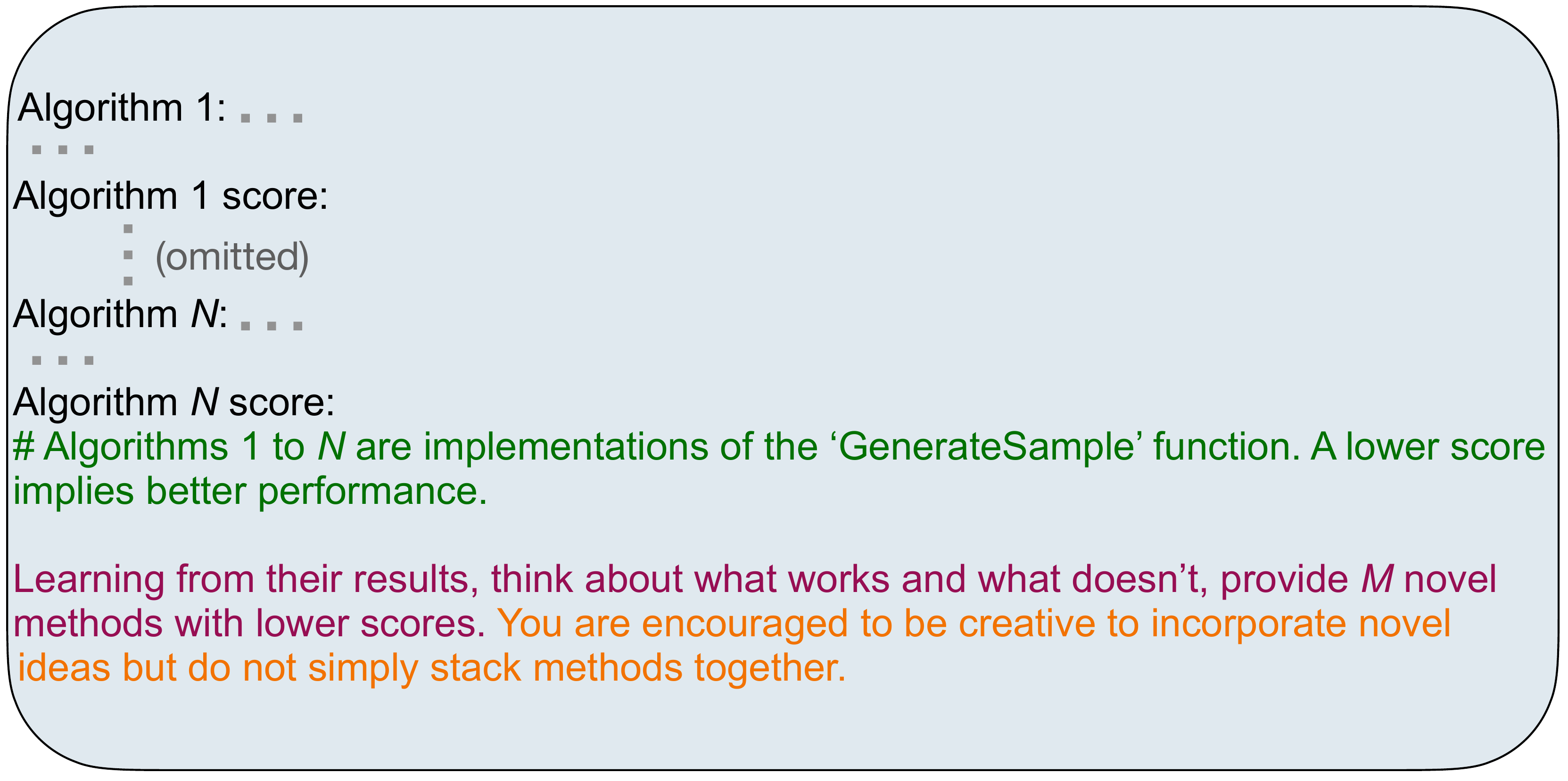}
    \vspace{-0.4cm}
    \caption{
The key prompt used in knowledge recombination (\texttt{KR}).
The \emph{\color{purple}purple} sentence specifies the goal and the \emph{\color{orange}orange} sentence clarifies the restriction.
    }
    \label{fig:KR}
\end{figure}

%% file: modules/II.tex
\begin{figure}[h]
    \centering
    \includegraphics[width=1\columnwidth]{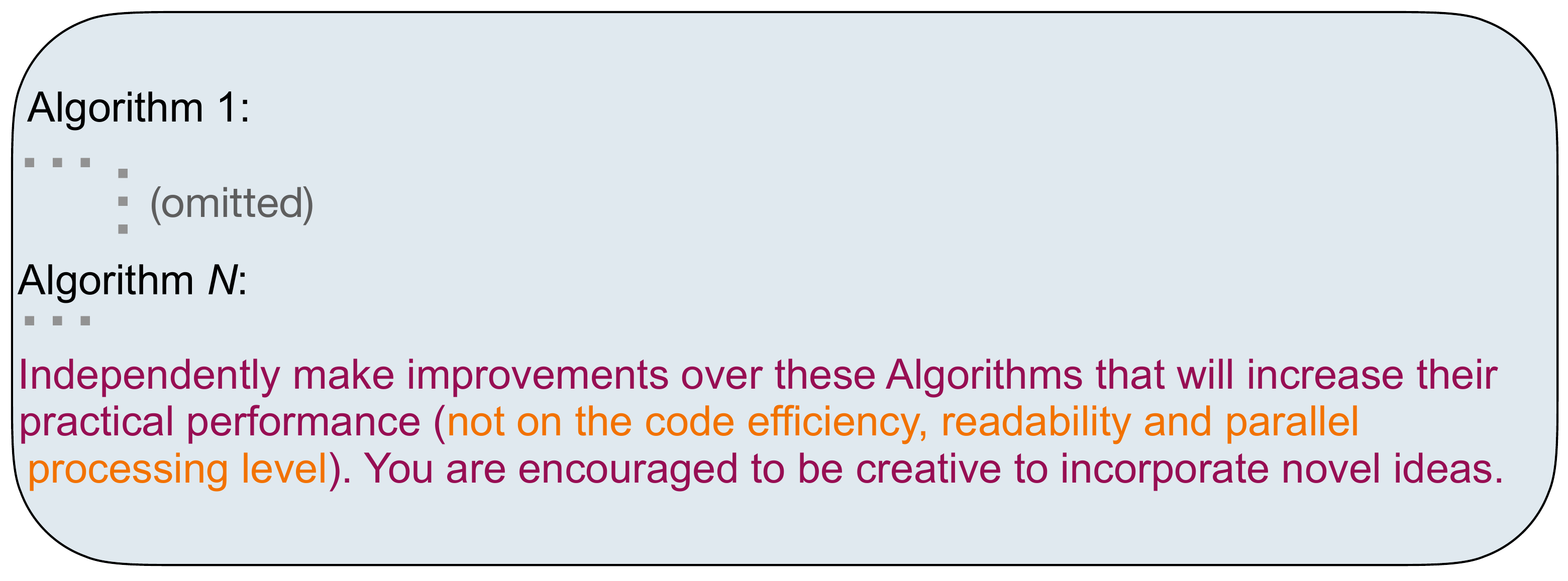}
    \vspace{-0.4cm}
    \caption{
The key prompt used in incremental innovation (\texttt{II}).
The \emph{\color{purple}purple} sentence specifies the goal and the \emph{\color{orange}orange} sentence clarifies the restriction.
    }
    \label{fig:II}
\end{figure}

%% file: modules/DI.tex
\begin{figure}[h]
    \centering
    \includegraphics[width=1\columnwidth]{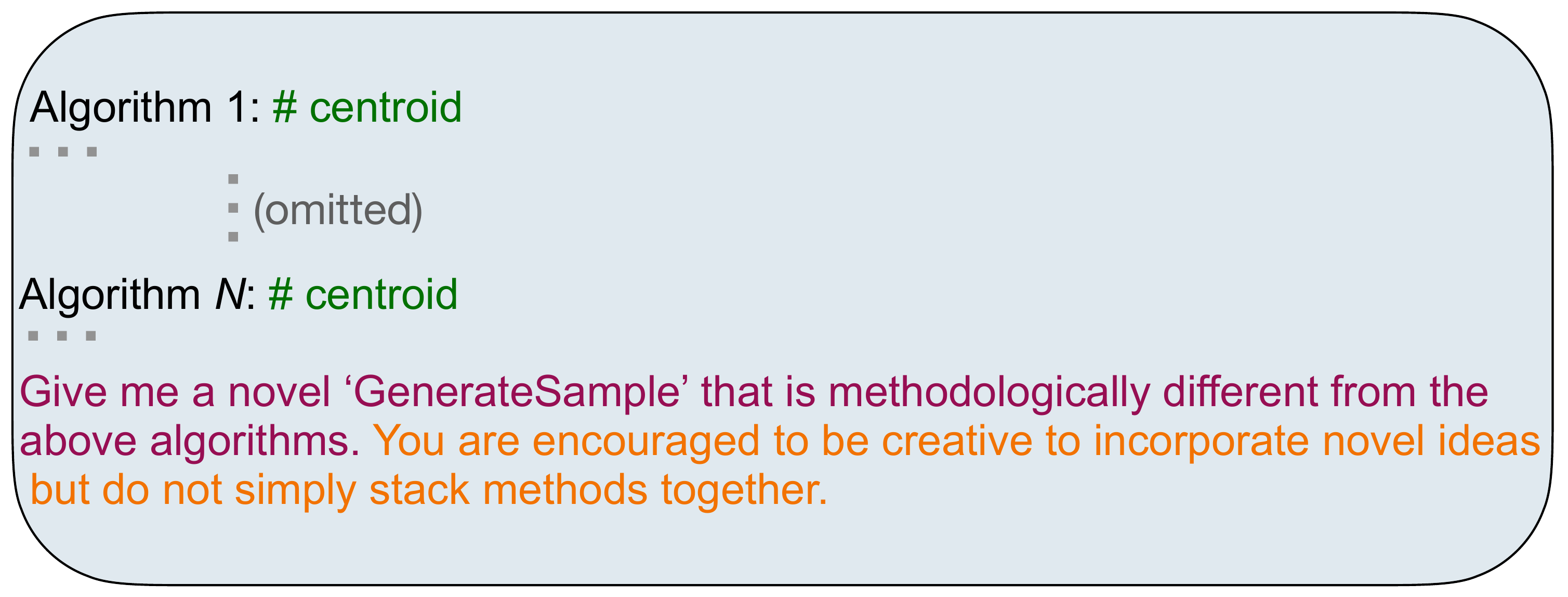}
    \vspace{-0.4cm}
    \caption{
The key prompt used in diversity injection (\texttt{DI}).
The \emph{\color{purple}purple} sentence specifies the goal and the \emph{\color{orange}orange} sentence clarifies the restriction.
    }
    \label{fig:DI}
\end{figure}

%% file: modules/FR.tex
\begin{figure}[h]
    \centering
    \includegraphics[width=1\columnwidth]{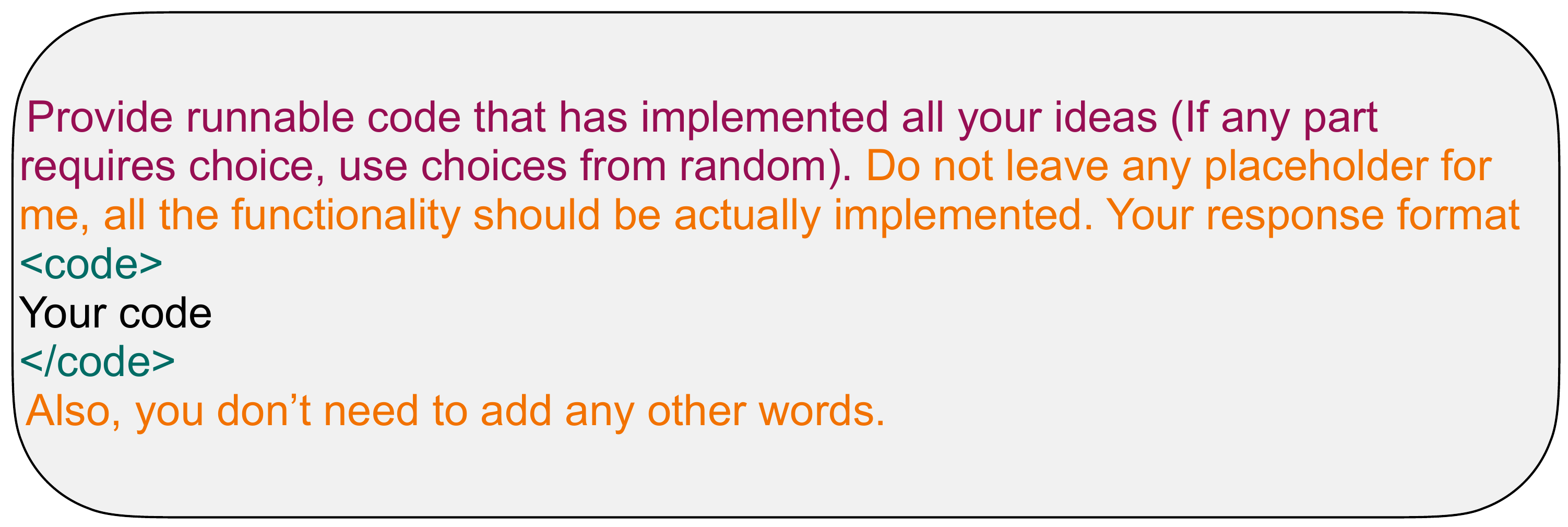}
    \vspace{-0.4cm}
    \caption{
The prompt used for format restriction of the output.
The \emph{\color{purple}purple} sentence specifies the goal and the \emph{\color{orange}orange} sentence clarifies the restriction.
    }
    \label{fig:FR}
\end{figure}

%% file: modules/frame.tex
\begin{figure}[h]
    \centering
    \includegraphics[width=1\columnwidth]{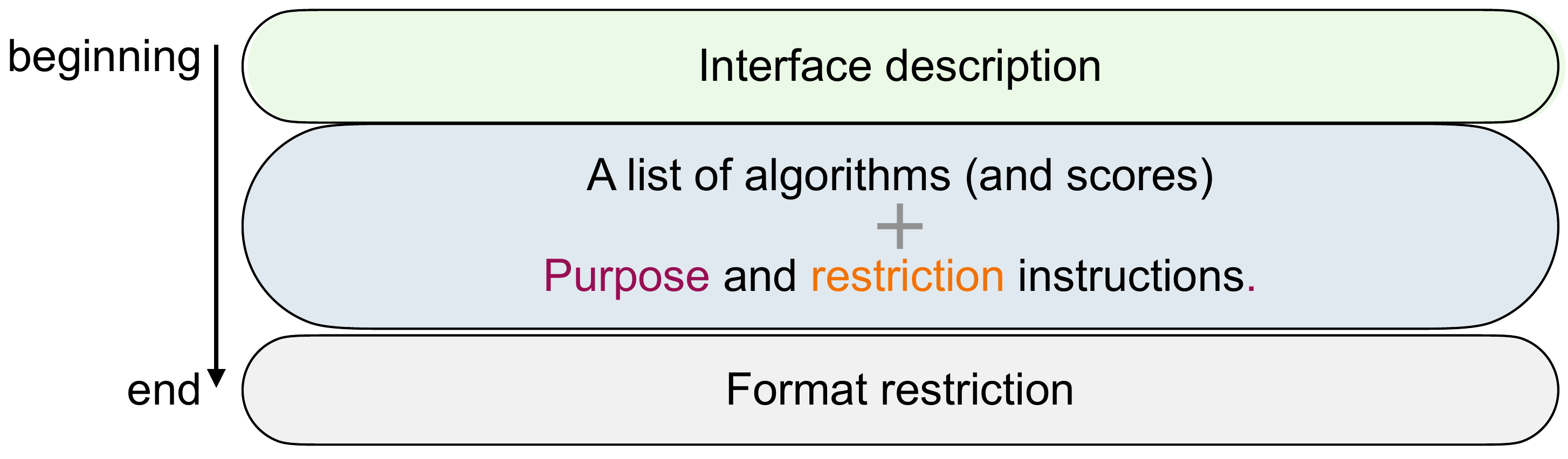}
    \vspace{-0.4cm}
    \caption{
Illustration to prompt architecture.
    }
    \label{fig:frame}
\end{figure}

%% file: modules/para.tex
\begin{table}[htbp]
  \centering
  \caption{Parameters involved in tnGPS.}
  \vspace{0.3cm}
  \label{tab:para}
  \begin{tabular}{lp{0.7\linewidth}}
    \hline
    \textbf{Parameter} & \textbf{Description} \\
    \hline
    $\#Iter$ & Maximum iteration \\
    $\alpha_1$ & Parameter in Eq.~\ref{eq:sampling} for cluster selection \\
    $\alpha_2$ & Parameter in Eq.~\ref{eq:sampling} for algorithm selection \\
    $m$ & Number of selected algorithms in \texttt{ID} \\
    $n$ & Number of generated algorithms \\
    $c$ & Maximum number of clusters \\
    \hline
  \end{tabular}
\end{table}

%% file: sections/99_appendix.tex
\begin{figure*}
    \centering
\includegraphics[width=1\columnwidth]{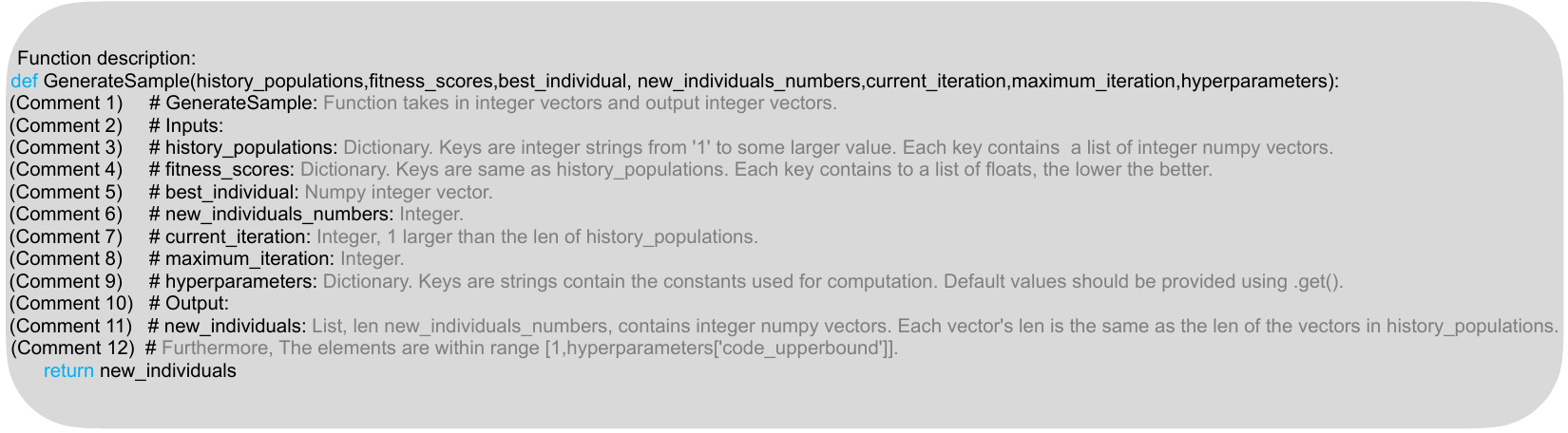}
    \caption{
Illustration of the interface description. In the interface description ablation experiment, comments $\{1,2,3,4,6,7,8,11\}$ are removed.
    }
    \label{fig:comments_removal}
\end{figure*}

\begin{figure*}[t!]
    \centering
\includegraphics[width=1\columnwidth]{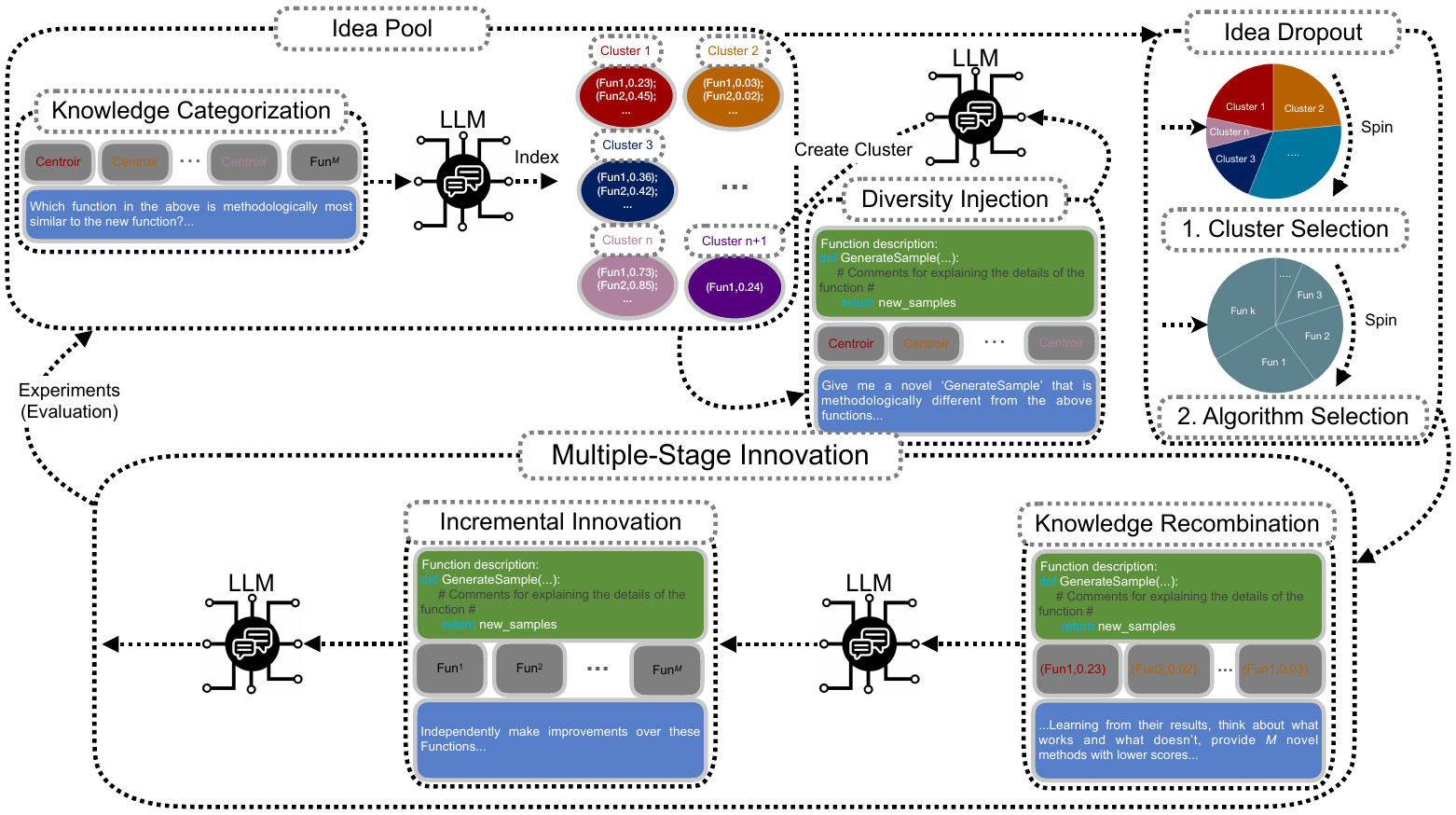}
    \caption{
The workflow of tnGPS. Each prompt can be constructed using up to three components: \textcolor[RGB]{94, 145, 60}{the interface description}, \textcolor[RGB]{127, 127, 127}{the in-context algorithms}, and \textcolor[RGB]{86, 126, 201}{the meta-prompt}. \textcolor[RGB]{94, 145, 60}{The interface description} is used to provide a precise objective, ensuring that the codes generated by the LLM are correct. \textcolor[RGB]{127, 127, 127}{The in-context algorithms} provide detailed information on the existing algorithms for LLM. Finally, \textcolor[RGB]{86, 126, 201}{the meta-prompt} is designed to guide the LLM. The other colors in the figure are utilized to indicate different clusters of algorithms. 
    }
    \label{fig:tnGPS_workflow}
\end{figure*}

\section{The workflow of tnGPS}\label{appendix:workflow}
In Figure \ref{fig:tnGPS_workflow}, we demonstrate the detailed workflow of tnGPS.


\section{Codes for different TN-SS algorithms}\label{appendix:code}
In the following, we present the codes for Ho-1, Ho-2, and Ho-3, as well as the codes of the algorithms discovered by tnGPS in the ablation experiments, and the codes for TNGA, TNLS, and GREEDY.

\subsection{Code for Ho-1}
\begin{minted}[fontsize=\scriptsize,breaklines]{python}
def GenerateSample(history_populations,fitness_scores,best_individual,new_individuals_numbers,current_iteration, maximum_iteration,hyperparameters):
    # Define default hyperparameters using .get()
    hyperparams = {
        'code_upperbound': hyperparameters.get('code_upperbound', 10),
        'mutation_rate': hyperparameters.get('mutation_rate', 0.15),
        'crossover_rate': hyperparameters.get('crossover_rate', 0.7),
        'selection_pressure': hyperparameters.get('selection_pressure', 1.8),
        'elitism_count': hyperparameters.get('elitism_count', 1),
        'mutation_scaling_factor': hyperparameters.get('mutation_scaling_factor', 0.9),
        'max_mutation': hyperparameters.get('max_mutation', 2),
        'tournament_size_factor': hyperparameters.get('tournament_size_factor', 0.15)
    }

    # Nested functions under GenerateSample
    def mutate(individual, scaling_factor):
        mutation_count = max(1, int(len(individual) * scaling_factor * hyperparams['mutation_rate']))
        mutation_indices = random.sample(range(len(individual)), mutation_count)
        for i in mutation_indices:
            individual[i] = random.randint(1, hyperparams['code_upperbound'])
        return individual

    def crossover(parent1, parent2):
        crossover_indices = random.sample(range(len(parent1)), int(len(parent1) * hyperparams['crossover_rate']))
        child = np.array([parent2[i] if i in crossover_indices else parent1[i] for i in range(len(parent1))])
        return child

    def tournament_selection(population, scores):
        tournament_size = max(int(len(population) * hyperparams['tournament_size_factor']), 2)
        selected_indices = random.sample(range(len(population)), tournament_size)
        selected_scores = [(i, scores[i]) for i in selected_indices]
        selected_scores.sort(key=lambda x: x[1])
        return population[selected_scores[0][0]]

    def create_new_individual(population, scores):
        parent1 = tournament_selection(population, scores)
        parent2 = tournament_selection(population, scores)
        child = crossover(parent1, parent2)
        mutation_scaling_factor = hyperparams['mutation_scaling_factor'] ** (1 + (maximum_iteration - current_iteration) / maximum_iteration)
        child = mutate(child, mutation_scaling_factor)
        return child

    def elitism(population, scores):
        sorted_population = sorted(zip(population, scores), key=lambda x: x[1])
        return [ind for ind, _ in sorted_population[:hyperparams['elitism_count']]]

    # Main logic for GenerateSample
    population = [np.array(individual) for key in sorted(history_populations.keys()) for individual in history_populations[key]]
    scores = [score for key in sorted(fitness_scores.keys()) for score in fitness_scores[key]]

    new_individuals = []
    elite_individuals = elitism(population, scores) if len(population) > 0 else []

    for elite in elite_individuals:
        new_individuals.append(elite)

    remaining_individuals_count = new_individuals_numbers - len(new_individuals)
    for _ in range(remaining_individuals_count):
        if len(population) > 0:
            new_individual = create_new_individual(population, scores)
        else:
            new_individual = np.random.randint(1, hyperparams['code_upperbound'] + 1, len(best_individual))
        new_individuals.append(new_individual)

    return new_individuals
\end{minted}
\newpage
\subsection{Code for Ho-2}
\begin{minted}[fontsize=\scriptsize,breaklines]{python}
def GenerateSample(history_populations,fitness_scores,best_individual,new_individuals_numbers,current_iteration, maximum_iteration,hyperparameters):
    hyperparams = {
        'code_upperbound': hyperparameters.get('code_upperbound', 10),
        'mutation_rate': hyperparameters.get('mutation_rate', 0.1),
        'crossover_rate': hyperparameters.get('crossover_rate', 0.6),
        'selection_pressure': hyperparameters.get('selection_pressure', 1.5),
        'elitism': hyperparameters.get('elitism', True),
        'diversity_factor': hyperparameters.get('diversity_factor', 0.05),
        'variance_decay': hyperparameters.get('variance_decay', 0.98),
        'variance_min': hyperparameters.get('variance_min', 0.1),
        'tournament_size_factor': hyperparameters.get('tournament_size_factor', 0.2),
        'elite_diversity_boost': hyperparameters.get('elite_diversity_boost', 2.0),
        'random_individual_chance': hyperparameters.get('random_individual_chance', 0.05),
        'max_mutation': hyperparameters.get('max_mutation', 3)
    }

    # Calculate variance based on current iteration
    variance = max(hyperparams['variance_decay'] ** (current_iteration - 1), hyperparams['variance_min'])

    def mutate(individual):
        # Perform mutation on an individual with limited number of gene changes
        mutation_indices = random.sample(range(len(individual)), min(len(individual), hyperparams['max_mutation']))
        for i in mutation_indices:
            if random.random() < hyperparams['mutation_rate']:
                individual[i] = random.randint(1, hyperparams['code_upperbound'])
        return individual

    def crossover(parent1, parent2):
        # Perform uniform crossover between two parents
        child = np.array([parent1[i] if random.random() < 0.5 else parent2[i] for i in range(len(parent1))])
        return child

    def select_parent(population, scores):
        # Perform tournament selection
        tournament_size = int(len(population) * hyperparams['tournament_size_factor'])
        selected_indices = random.sample(range(len(population)), tournament_size)
        selected_scores = [scores[i] for i in selected_indices]
        winner_index = selected_indices[selected_scores.index(min(selected_scores))]
        return population[winner_index]

    def introduce_diversity(individual, diversity_boost=1.0):
        # Introduce diversity to an individual by adding Gaussian noise
        noise = np.random.randn(len(individual)) * variance * diversity_boost
        individual = np.round(individual + noise)
        individual = np.clip(individual, 1, hyperparams['code_upperbound']).astype(int)
        return individual

    def create_random_individual(length):
        return np.random.randint(1, hyperparams['code_upperbound'] + 1, length)

    # Convert history_population to a list of numpy arrays and fitness_scores to a list of scores
    population = [np.array(individual) for key in sorted(history_populations.keys()) for individual in history_populations[key]]
    scores = [score for key in sorted(fitness_scores.keys()) for score in fitness_scores[key]]

    # Generate new individuals
    new_individuals = []
    for _ in range(new_individuals_numbers):
        if random.random() < hyperparams['random_individual_chance']:
            new_individual = create_random_individual(len(best_individual))
        elif hyperparams['elitism'] and best_individual is not None and random.random() < hyperparams['diversity_factor']:
            # Add a mutated and diversified version of the best individual
            new_individual = mutate(introduce_diversity(best_individual.copy(), hyperparams['elite_diversity_boost']))
        elif len(population) > 0:
            # Create a new individual using crossover and mutation
            parent1 = select_parent(population, scores)
            parent2 = select_parent(population, scores)
            child = crossover(parent1, parent2)
            child = mutate(child)
            new_individual = introduce_diversity(child)
        else:
            # If there is no history population, create a random individual
            new_individual = create_random_individual(len(best_individual))
        new_individuals.append(new_individual)

    return new_individuals
\end{minted}
\newpage
\subsection{Code for Ho-3}
\begin{minted}[fontsize=\scriptsize,breaklines]{python}
def GenerateSample(history_populations,fitness_scores,best_individual,new_individuals_numbers,current_iteration, maximum_iteration,hyperparameters):
    # Define default hyperparameters
    hyperparams = {
        'code_upperbound': hyperparameters.get('code_upperbound', 10),
        'mutation_rate': hyperparameters.get('mutation_rate', 0.2),
        'crossover_rate': hyperparameters.get('crossover_rate', 0.5),
        'selection_pressure': hyperparameters.get('selection_pressure', 2.0),
        'elitism': hyperparameters.get('elitism', True),
    }

    def mutate(individual):
        # Perform mutation on an individual
        for i in range(len(individual)):
            if random.random() < hyperparams['mutation_rate']:
                individual[i] = random.randint(1, hyperparams['code_upperbound'])
        return individual

    def crossover(parent1, parent2):
        # Perform crossover between two parents
        child = parent1.copy()
        for i in range(len(child)):
            if random.random() < hyperparams['crossover_rate']:
                child[i] = parent2[i]
        return child

    def select_parent(population, scores):
        # Perform tournament selection
        tournament_size = min(len(population), int(len(population) * hyperparams['selection_pressure']))
        selected_indices = random.sample(range(len(population)), tournament_size)
        selected_scores = [scores[i] for i in selected_indices]
        winner_index = selected_indices[selected_scores.index(min(selected_scores))]
        return population[winner_index]

    def create_new_individual(population, scores):
        # Create a new individual using crossover and mutation
        if hyperparams['elitism'] and best_individual is not None:
            parent1 = best_individual
        else:
            parent1 = select_parent(population, scores)
        
        parent2 = select_parent(population, scores)
        child = crossover(parent1, parent2)
        child = mutate(child)
        return child

    # Convert history_population to a list of numpy arrays and fitness_scores to a list of scores
    population = [np.array(individual) for key in sorted(history_populations.keys()) for individual in history_populations[key]]
    scores = [score for key in sorted(fitness_scores.keys()) for score in fitness_scores[key]]
    
    # Generate new individuals
    new_individuals = []
    for _ in range(new_individuals_numbers):
        if len(population) > 0:
            new_individual = create_new_individual(population, scores)
        else:
            # If there is no history population, create a random individual
            new_individual = np.random.randint(1, hyperparams['code_upperbound'] + 1, len(best_individual))
        new_individuals.append(new_individual)
    
    return new_individuals
\end{minted}

\subsection{Code of the algorithm discovered by tnGPS in the tnGPS components ablation experiment}
\begin{minted}[fontsize=\scriptsize,breaklines]{python}
def GenerateSample(history_populations, fitness_scores, best_individual, new_individuals_numbers, current_iteration, maximum_iteration, hyperparameters):
    
    def tournament_selection(populations, fitness_scores, tournament_size):
        selected_indices = []
        for _ in range(tournament_size):
            participants = choices(range(len(populations)), k=tournament_size)
            participants_fitness = [fitness_scores[i] for i in participants]
            winner_index = participants[np.argmin(participants_fitness)]
            selected_indices.append(winner_index)
        return [populations[i] for i in selected_indices]

    def uniform_crossover(parent1, parent2, crossover_rate):
        child = np.array([p1 if random() < crossover_rate else p2 for p1, p2 in zip(parent1, parent2)])
        return child

    def boundary_mutation(individual, mutation_rate, code_upperbound):
        for i in range(len(individual)):
            if random() < mutation_rate:
                individual[i] = 1 if random() < 0.5 else code_upperbound
        return individual

    # Retrieve hyperparameters with defaults
    tournament_size = hyperparameters.get('tournament_size', 3)
    crossover_rate = hyperparameters.get('crossover_rate', 0.7)  # Increased crossover rate for potentially better offspring
    mutation_rate = hyperparameters.get('mutation_rate', 0.05)  # Reduced mutation rate to maintain good traits
    code_upperbound = hyperparameters.get('code_upperbound', 100)
    elitism_count = hyperparameters.get('elitism_count', 1)  # Introducing elitism to ensure the best individual is carried forward

    current_population = history_populations[str(current_iteration - 1)]
    current_fitness = fitness_scores[str(current_iteration - 1)]

    # Sort the current population by fitness and apply elitism
    sorted_indices = np.argsort(current_fitness)
    elites = [current_population[i] for i in sorted_indices[:elitism_count]]

    # Generate new individuals, starting with the elites
    new_individuals = elites.copy()
    while len(new_individuals) < new_individuals_numbers:
        # Tournament selection
        parents = tournament_selection(current_population, current_fitness, tournament_size)
        # Uniform crossover
        child1 = uniform_crossover(parents[0], parents[1], crossover_rate)
        child2 = uniform_crossover(parents[1], parents[0], crossover_rate)
        # Boundary mutation
        child1 = boundary_mutation(child1, mutation_rate, code_upperbound)
        child2 = boundary_mutation(child2, mutation_rate, code_upperbound)
        # Add children to the new population
        new_individuals.extend([child1, child2])
    
    # Truncate in case we have extra individuals
    return new_individuals[:new_individuals_numbers]
\end{minted}

\subsection{Code of the algorithm discovered by tnGPS in the LLM models and interface description ablation experiment}
\begin{minted}[fontsize=\scriptsize,breaklines]{python}
def GenerateSample(history_populations, fitness_scores, best_individual, new_individuals_numbers, current_iteration, maximum_iteration, hyperparameters):
    
    # Hyperparameters with default values
    hyper = {
        'code_upperbound': hyperparameters.get('code_upperbound', 30),
        'mutation_rate': hyperparameters.get('mutation_rate', 0.1),
        'tournament_size': hyperparameters.get('tournament_size', 5),
        'elitism_rate': hyperparameters.get('elitism_rate', 0.1),
        'crossover_rate': hyperparameters.get('crossover_rate', 0.9),
        'diversity_factor': hyperparameters.get('diversity_factor', 0.1),
        'best_individual_influence': hyperparameters.get('best_individual_influence', 0.05)
    }
    
    def tournament_selection(populations, fitness_scores, tournament_size):
        tournament_contestants = choices(list(zip(populations, fitness_scores)), k=tournament_size)
        tournament_contestants.sort(key=lambda x: x[1])  # sort by fitness score, lower is better
        winner = tournament_contestants[0][0]  # return the individual with the best fitness
        return winner

    def mutate(individual, mutation_rate, code_upperbound):
        mutation_indices = [i for i in range(len(individual)) if random() < mutation_rate]
        for i in mutation_indices:
            individual[i] = randint(1, code_upperbound)
        return individual

    def crossover(parent1, parent2, crossover_rate, best_individual, best_influence):
        child = parent1.copy()
        if random() < crossover_rate:
            for i in range(len(parent1)):
                if random() < best_influence:
                    child[i] = best_individual[i]
                elif random() < 0.5:
                    child[i] = parent1[i]
                else:
                    child[i] = parent2[i]
        return child

    def introduce_diversity(population, diversity_factor, code_upperbound):
        for individual in population:
            if random() < diversity_factor:
                mutation_index = randint(0, len(individual) - 1)
                individual[mutation_index] = randint(1, code_upperbound)
        return population

    # Retrieve the latest population and their fitness scores
    elite_population = history_populations[str(current_iteration - 1)]
    elite_fitness = fitness_scores[str(current_iteration - 1)]

    # Calculate the number of elites based on the elitism rate
    number_of_elites = int(hyper['elitism_rate'] * new_individuals_numbers)

    # Sort the elite_population based on fitness and select the top individuals
    sorted_indices = np.argsort(elite_fitness)
    elites = [elite_population[i] for i in sorted_indices[:number_of_elites]]

    # Generate new individuals with crossover and mutation
    new_individuals = []
    while len(new_individuals) < new_individuals_numbers - number_of_elites:
        parent1 = tournament_selection(elite_population, elite_fitness, hyper['tournament_size'])
        parent2 = tournament_selection(elite_population, elite_fitness, hyper['tournament_size'])
        child = crossover(parent1, parent2, hyper['crossover_rate'], best_individual, hyper['best_individual_influence'])
        new_individuals.append(mutate(child, hyper['mutation_rate'], hyper['code_upperbound']))

    # Introduce diversity
    new_individuals = introduce_diversity(new_individuals, hyper['diversity_factor'], hyper['code_upperbound'])

    # Include elites in the new population pool
    new_individuals.extend(elites)

    # Ensure all values are within the specified range
    new_individuals = [np.clip(individual, 1, hyper['code_upperbound']) for individual in new_individuals]

    return new_individuals
\end{minted}

\subsection{Code for TNGA}
\begin{minted}[fontsize=\scriptsize,breaklines]{python}
def GenerateSample(history_populations,fitness_scores,best_individual,new_individuals_numbers,current_iteration, maximum_iteration,hyperparameters):
    Ranking=np.argsort(fitness_scores['{}'.format(current_iteration-1)])
    elite_num=int(len(fitness_scores['{}'.format(current_iteration-1)])*hyperparameters.get('elite_percentage', 0.9))
    Ranking=Ranking[0:elite_num]
    populations_elite=[history_populations['{}'.format(current_iteration-1)][i].copy() for i in Ranking]
    fitness_scores_elite=[fitness_scores['{}'.format(current_iteration-1)][i] for i in Ranking]
    Rank_elite = np.argsort(fitness_scores_elite)
    p = [ np.maximum(np.log(hyperparameters.get('alpha', 100)/(0.01+k*5)), 0.01) for k in range(len(populations_elite)) ]
    prob = np.zeros(len(populations_elite))
    for idx, i in enumerate(Rank_elite): prob[i] = p[idx]
    new_individuals=[]
    for i in range(new_individuals_numbers//2): 
        parents=choices(populations_elite, weights=prob, k=2)
        female=parents[0].copy()
        male=parents[1].copy()
        index=np.arange(len(male))
        np.random.shuffle(index)
        index=index[0:(len(male)//2)]
        tnp=female[index]
        female[index]=male[index]
        male[index]=tnp
        new_individuals.append(male)
        new_individuals.append(female)
    if np.mod(new_individuals_numbers,2)!=0:
        tnp=new_individuals[-1].copy()
        np.random.shuffle(tnp)
        new_individuals.append(tnp)
    for i in range(new_individuals_numbers):
        mask = np.random.uniform(0,1,[len(new_individuals[0])])<hyperparameters.get('mutation_rate', 0.25)
        for j in range(len(new_individuals[0])):
            if mask[j]:
                mutate_range=np.arange(1,hyperparameters.get('code_upperbound', 15)+1)
                mutate_range=np.delete(mutate_range, np.where(mutate_range == new_individuals[i][j]))
                np.random.shuffle(mutate_range)
                new_individuals[i][j]=mutate_range[0]
    return new_individuals
\end{minted}
\subsection{Code for TNLS}
\begin{minted}[fontsize=\scriptsize,breaklines]{python}
def GenerateSample(history_populations,fitness_scores,best_individual,new_individuals_numbers,current_iteration, maximum_iteration,hyperparameters):
     variance=hyperparameters.get('decay_rate', 0.99)**(current_iteration-2)
     if variance<hyperparameters.get('variance_LB', 0.3):
         variance=hyperparameters.get('variance_LB', 0.3)
     new_individuals=[]
     for i in range(new_individuals_numbers):
         tnp=np.array(best_individual)+np.random.randn(len(best_individual))*variance
         tnp=np.round(tnp)
         tnp[np.where(tnp>hyperparameters.get('code_upperbound', 15))]=hyperparameters.get('code_upperbound', 15)
         tnp[np.where(tnp<1)]=1
         tnp=tnp.astype(int)
         new_individuals.append(tnp)
     return new_individuals
\end{minted}
\subsection{Code for GREEDY}
\begin{minted}[fontsize=\scriptsize,breaklines]{python}
def GenerateSample(history_populations,fitness_scores,best_individual,new_individuals_numbers,current_iteration, maximum_iteration,hyperparameters):
     variance=hyperparameters.get('decay_rate', 0.99)**(current_iteration-2)
     if variance<hyperparameters.get('variance_LB', 0.3):
         variance=hyperparameters.get('variance_LB', 0.3)
     new_individuals=[]
     for i in range(new_individuals_numbers):
         tnp=np.array(best_individual)+np.random.randn(len(best_individual))*variance
         tnp=np.round(tnp)
         tnp[np.where(tnp>hyperparameters.get('code_upperbound', 15))]=hyperparameters.get('code_upperbound', 15)
         tnp[np.where(tnp<1)]=1
         tnp=tnp.astype(int)
         new_individuals.append(tnp)
     return new_individuals
\end{minted}
